\documentclass[11pt]{article}

\usepackage[preprint]{acl}

\usepackage{times}
\usepackage{latexsym}

\usepackage[T1]{fontenc}

\usepackage[utf8]{inputenc}

\usepackage{microtype}

\usepackage{inconsolata}

\usepackage{graphicx}

\usepackage{hyperref}
\usepackage{url}
\usepackage{subcaption}

\usepackage[utf8]{inputenc}
\usepackage[T1]{fontenc}
\usepackage{hyphenat}
\usepackage{xspace}
\usepackage{amsmath}
\usepackage{amsfonts}
\usepackage{hyperref}
\usepackage{url}
\usepackage{booktabs}
\usepackage{multirow}
\usepackage{makecell}
\usepackage{caption}
\usepackage{minibox}
\usepackage{bbm}
\usepackage{graphicx}
\usepackage{balance}
\usepackage{mathtools}
\usepackage{color}
\usepackage{marvosym}
\usepackage{ifthen}
\usepackage{textcomp}
\usepackage{enumitem}
\usepackage{verbatim}
\usepackage{algorithm}
\usepackage{numprint}
\usepackage{balance}

\usepackage{amsthm}
\theoremstyle{plain}

\newcommand{\chatoDisplayMode}[1]{#1}

\definecolor{MyRed}{rgb}{0.6,0.0,0.0} 
\definecolor{MyBlack}{rgb}{0.1,0.1,0.1} 
\newcommand{\inred}[1]{{\color{MyRed}\sf\textbf{\textsc{#1}}}}
\newcommand{\frameit}[2]{
  \begin{center}
  {\color{MyRed}
  \framebox[.9\columnwidth][l]{
    \begin{minipage}{.85\columnwidth}
    \inred{#1}: {\sf\color{MyBlack}#2}
    \end{minipage}
  }\\
  }
  \end{center}
}

\newcommand{\note}[2][]{\chatoDisplayMode{\def\@tmpsig{#1}\frameit{{\Pointinghand} Note}{#2\ifx \@tmpsig \@empty \else \mbox{ --\em #1}\fi}}}
\newcommand{\todo}[2][]{\chatoDisplayMode{\def\@tmpsig{#1}\frameit{{\Writinghand} To-do}{#2\ifx \@tmpsig \@empty \else \mbox{ --\em #1}\fi}}}

\newcommand{\abbrevStyle}[1]{#1}

\newcommand{\ie}{\abbrevStyle{i.e.}\xspace}
\newcommand{\eg}{\abbrevStyle{e.g.}\xspace}
\newcommand{\cf}{\abbrevStyle{cf.}\xspace}

\newcommand{\vs}{\abbrevStyle{vs.}\xspace}

\newcommand{\Secref}[1]{Sec.~\ref{#1}}

\newcommand{\Figref}[1]{Fig.~\ref{#1}}

\newcommand{\Appref}[1]{Appendix~\ref{#1}}

\newcommand{\xhdr}[1]{\vspace{1.7mm}\noindent{{\bf #1.}}}

\newcommand{\denselist}{ \itemsep -2pt\topsep-10pt\partopsep-10pt }

\newcommand{\textcite}[1]{\citeauthor{#1} \shortcite{#1}}

\newcommand{\hide}[1]{}

\makeatletter
\newcommand{\iffont}[2]{\ifthenelse{\equal{\f@family}{#1}}{#2}{}}
\makeatother

  \usepackage{mathptmx}

  \DeclareSymbolFont{greek}{OML}{cmm}{m}{n}
  \DeclareMathSymbol{\alpha}{\mathalpha}{greek}{"0B}
  \DeclareMathSymbol{\beta}{\mathalpha}{greek}{"0C}
  \DeclareMathSymbol{\gamma}{\mathalpha}{greek}{"0D}
  \DeclareMathSymbol{\delta}{\mathalpha}{greek}{"0E}
  \DeclareMathSymbol{\epsilon}{\mathalpha}{greek}{"0F}
  \DeclareMathSymbol{\zeta}{\mathalpha}{greek}{"10}
  \DeclareMathSymbol{\eta}{\mathalpha}{greek}{"11}
  \DeclareMathSymbol{\theta}{\mathalpha}{greek}{"12}
  \DeclareMathSymbol{\iota}{\mathalpha}{greek}{"13}
  \DeclareMathSymbol{\kappa}{\mathalpha}{greek}{"14}
  \DeclareMathSymbol{\lambda}{\mathalpha}{greek}{"15}
  \DeclareMathSymbol{\mu}{\mathalpha}{greek}{"16}
  \DeclareMathSymbol{\nu}{\mathalpha}{greek}{"17}
  \DeclareMathSymbol{\xi}{\mathalpha}{greek}{"18}
  \DeclareMathSymbol{\pi}{\mathalpha}{greek}{"19}
  \DeclareMathSymbol{\rho}{\mathalpha}{greek}{"1A}
  \DeclareMathSymbol{\sigma}{\mathalpha}{greek}{"1B}
  \DeclareMathSymbol{\tau}{\mathalpha}{greek}{"1C}
  \DeclareMathSymbol{\upsilon}{\mathalpha}{greek}{"1D}
  \DeclareMathSymbol{\phi}{\mathalpha}{greek}{"1E}
  \DeclareMathSymbol{\chi}{\mathalpha}{greek}{"1F}
  \DeclareMathSymbol{\psi}{\mathalpha}{greek}{"20}
  \DeclareMathSymbol{\omega}{\mathalpha}{greek}{"21}
  \DeclareMathSymbol{\varepsilon}{\mathalpha}{greek}{"22}
  \DeclareMathSymbol{\vartheta}{\mathalpha}{greek}{"23}
  \DeclareMathSymbol{\varpi}{\mathalpha}{greek}{"24}
  \DeclareMathSymbol{\varrho}{\mathalpha}{greek}{"25}
  \DeclareMathSymbol{\varsigma}{\mathalpha}{greek}{"26}
  \DeclareMathSymbol{\varphi}{\mathalpha}{greek}{"27}
  \DeclareSymbolFont{otone}{OT1}{cmr}{m}{n}
  \DeclareMathSymbol{\Gamma}{\mathalpha}{otone}{0}
  \DeclareMathSymbol{\Delta}{\mathalpha}{otone}{1}
  \DeclareMathSymbol{\Theta}{\mathalpha}{otone}{2}
  \DeclareMathSymbol{\Lambda}{\mathalpha}{otone}{3}
  \DeclareMathSymbol{\Xi}{\mathalpha}{otone}{4}
  \DeclareMathSymbol{\Pi}{\mathalpha}{otone}{5}
  \DeclareMathSymbol{\Sigma}{\mathalpha}{otone}{6}
  \DeclareMathSymbol{\Upsilon}{\mathalpha}{otone}{7}
  \DeclareMathSymbol{\Phi}{\mathalpha}{otone}{8}
  \DeclareMathSymbol{\Psi}{\mathalpha}{otone}{9}
  \DeclareMathSymbol{\Omega}{\mathalpha}{otone}{10}
  \DeclareSymbolFont{syms}{OML}{cmm}{m}{it}
  \DeclareMathSymbol{\partial}{\mathord}{syms}{"40}
  \DeclareMathAlphabet{\mathbold}{OML}{cmm}{b}{it}
  \DeclareSymbolFont{largesymbols}{OMX}{cmex}{m}{n}

  \DeclareMathAlphabet{\mathcal}{OMS}{cmsy}{m}{n}

\newcommand{\lgui}{$\ll$}
\newcommand{\rgui}{$\gg$}

\newcommand{\sen}{\text{sen}}
\newcommand{\jun}{\text{jun}}

\newcommand{\Lsen}{L_\sen}
\newcommand{\Ljun}{L_\jun}
\newcommand{\Msen}{M_\sen}
\newcommand{\Mjun}{M_\jun}

\newcommand{\llama}{Llama-2\xspace}

\author{%
Robert West,\textsuperscript{1}\thanks{Work done as visiting researchers at Microsoft Research.}
Ashton Anderson,\textsuperscript{2\textnormal{\textasteriskcentered}}
Ece Kamar,\textsuperscript{3}
Eric Horvitz\textsuperscript{3}\\
\textsuperscript{1}EPFL,
\textsuperscript{2}University of Toronto,
\textsuperscript{3}Microsoft\\
{robert.west@epf\/l.ch},
{ashton@cs.toronto.edu},
{eckamar@microsoft.com},
{horvitz@microsoft.com}
\\
}

\title{Tandem Training for Language Models}

\begin{document}

\maketitle

\begin{abstract}
As language models continue to rapidly improve, we can expect their actions and reasoning to become difficult or impossible for weaker agents and humans to follow, undermining interpretability and oversight. With an eye on long-term futures, we pursue methods that encourage models to produce solutions that remain intelligible to weaker collaborators. We formalize intelligibility as \emph{handoff robustness:} a strong model's solution is intelligible to a weaker model if randomly handing off control to the weaker model along the solution path does not cause failure. Building on this criterion, we introduce \emph{tandem training} for language models, a reinforcement learning (RL) paradigm in which rollout tokens are intermittently and randomly sampled from a frozen weak model rather than the strong model being trained. Because rollouts succeed only when the strong model's actions and reasoning process can be continued by the weak model---when the two can co-construct a successful solution---optimizing standard RL objectives with tandem training implicitly incentivizes both correctness and intelligibility. In the GSM8K math reasoning task, tandem training reliably teaches models to abandon jargon and adapt their language to weaker partners while keeping task accuracy high. Our results demonstrate a promising route to building AI systems that remain auditable by weaker agents, with implications for human--AI collaboration and multi-agent communication.
\end{abstract}

\section{Introduction}
\label{sec:Introduction}

\begin{figure*}[t]
    \centering
    \includegraphics[width=\linewidth]{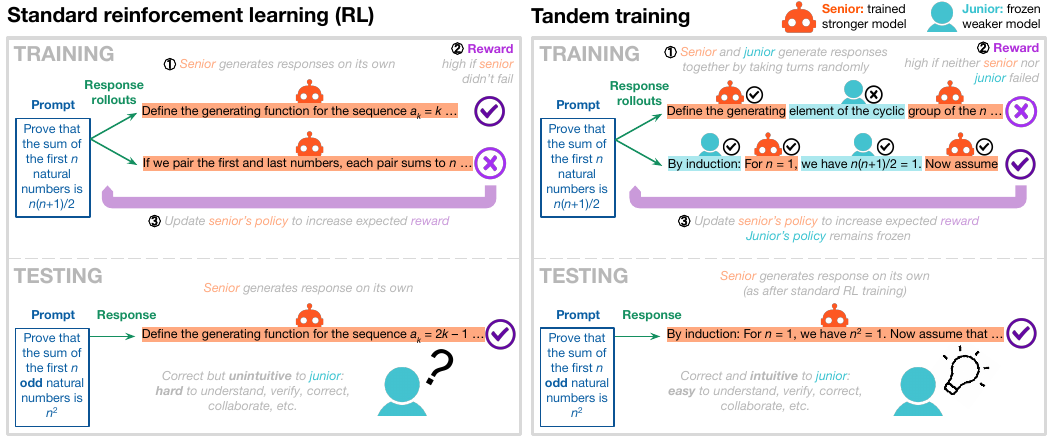}
    \caption{
\textbf{Tandem training \vs\ standard RL.}
Standard RL \textbf{\textit{(left)}} only encourages correct solutions, regardless of whether they are intelligible to other models or humans, which raises concerns about interpretability and oversight.
We define intelligibility via \textit{handoff robustness,} positing that a partial solution is intelligible to another agent if that agent could continue the solution, at least for a few steps, without derailing it into failure.
\textit{Tandem training} \textbf{\textit{(right)}} encourages handoff robustness, and thus intelligibility, by letting the trained model (``senior'') generate solutions together with a frozen (typically weaker) ``junior'' model, taking turns randomly. Since rewards are high only if neither senior nor junior made a critical mistake, the senior is encouraged to generate solutions that can be correctly continued (and thus, by our definition, understood) by the junior.
    }
    \label{fig:tandem training}
\end{figure*}

Artificial intelligence (AI) is rapidly becoming more capable. AI models have already matched or surpassed  human capabilities in several milestone domains, and most researchers expect this trend to continue. As AI improves, however, its actions and reasoning will often become difficult or impossible for weaker agents and humans to follow because AI improvement is often driven by autonomous learning loops, synthetic data, and proxy rewards, causing model behavior to potentially stray from what humans or other agents might find familiar or intuitive. 

This poses serious risks and challenges. Highly capable but unintelligible models may be useful in isolation or in highly modular environments. But when collaboration is useful or necessary, or when oversight is required, or in any scenario where interpretability and control are important, unintelligibility is a major problem. An agent with hard-to-follow reasoning, uninterpretable actions, or unpredictable decision-making increases the risk of detrimental outcomes in any system in which it plays a part. Without being intelligible, AI models and agents are limited in the extent to which they can collaborate with people, integrate into multi-agent systems, and be broadly beneficial.

With long-term futures in mind, we pursue methods that encourage models to produce high-quality solutions while remaining intelligible to weaker collaborators. We focus on language models, given their ubiquity and ability. One way to approach the problem of intelligibility would be to design elaborate system prompts, provide supervised examples, or construct custom rewards, but these bespoke solutions would be fraught with challenges. They would likely be brittle, easily gamed, and limited in scope.

We adopt another approach, where intelligibility is defined in a pragmatic, outcome-based manner: a (stronger) senior model $\Msen$ is considered intelligible to a (weaker) junior model $\Mjun$ if $\Mjun$ can take over during randomly assigned portions of a solution path without causing task failure. Imagine a model in the process of solving a problem, when randomly and unpredictably, it must hand off control to a weaker collaborator to continue solving the problem, and this handoff may occur several times on the solution path. If $\Msen$ and $\Mjun$ can jointly construct successful task solutions, then $\Mjun$ must be able to understand $\Msen$'s behavior and actions well enough to continue successfully. We consider this a strong form of intelligibility, as the weaker collaborator is not simply interpreting the stronger model's actions \textit{post hoc,} but actively co-constructs successful solutions.

Such \emph{handoff robustness} is not straightforward to achieve. We tackle the challenge by introducing a novel paradigm for training language models, which we call \emph{tandem training,} inspired by a proof of concept in chess~\citep{hamade2024designing}.
Tandem training (\cf\ \Figref{fig:tandem training}) can be combined with any reinforcement learning (RL) algorithm used for language model training. In RL, solution rollouts are sampled from the language model being trained, with the goal of training being to make successful rollouts more likely. Tandem training modifies the rollout phase by sometimes sampling next tokens from a frozen junior model $\Mjun$ rather than the senior model $\Msen$ being trained. Whether the next turn's tokens are sampled from the junior or the senior model is determined randomly. As rollouts are more likely to be successful when $\Msen$ generates text that can be continued by $\Mjun$ without introducing errors that lead to task failure, optimizing $\Msen$ via tandem training incentivizes not only correct, but also intelligible, solutions.

We conduct experiments in the domain of mathematical reasoning across three settings and find that tandem training reliably
teaches strong models to abandon jargon and adapt their language to weaker models while keeping task accuracy high.
In settings where senior and junior differ in their math skills, senior-specific notation drops from 99\% to 0\% within 20 gradient updates---indicating adaptation to the frozen junior model, which does not understand this notation---yet senior accuracy remains above the junior baseline.
In settings where senior and junior use different languages before training, the senior adapts its language to the junior's language, again without senior accuracy falling below the junior baseline.
These results demonstrate
a practical path to aligning capable models with weaker collaborators,
with the potential to
improve AI model intelligibility and enable safer, more reliable, and more performant AI--human and AI--AI interaction.

\section{Related work}
\label{sec:Related work}

\xhdr{Tandem training}
Our work applies tandem training, an RL paradigm first introduced by~\citet{hamade2024designing} in chess, to language models. Their work showed that optimizing for partner compatibility (winning a team chess game) is distinct from optimizing for raw ability (winning a chess game on one's own). We adopt their framework and setup (\eg, ``senior'' and ``junior'' models; randomized handoffs between them) and significantly expand on this work by applying tandem training to language models, solving reasoning problems, and moving beyond the adversarial game environment. 

\xhdr{RL with verifiable rewards}
Recent work demonstrates that pure RL with automatically verifiable outcome rewards
can produce strong reasoners~\citep{wen2025rlvr}, but with the undesirable pathology of degraded intelligibility~\citep{li2025impact,guo2025deepseek,wu2025more}. Reasoning chains often become less reliable, grounded, and interpretable with length and complexity~\citep{cheng2025chain,hassid2025don}.

\xhdr{Scalable oversight}
There has been a recent push to enable scalable oversight, where strong models can be overseen by weaker agents, \eg, via decomposition \citep{christiano2018supervising} or debate \citep{irving2018debate}. Complementary lines replace scarce human labels with AI feedback or process-level signals, showing that supervising intermediate reasoning can improve correctness and auditability at scale \citep{bai2022training,paul2023refiner,lightman2023verify}. Research on weak-to-strong generalization \citep{burns2024weak} demonstrates that training with weak supervision can nevertheless surface strong capabilities, formalizing conditions under which a weaker teacher or signal suffices. Our tandem training approach operationalizes scalable oversight inside the solution trajectory: instead of relying on external judges, decompositions, or teacher labels, we impose random handoffs to a weaker collaborator during RL and reward success only when the strong model's reasoning is continuable by weaker agents, incentivizing intelligible solutions that are more amenable to oversight.

\section{Method: tandem training}
\label{sec:Method}

There are many domains of practical interest in which we would prefer language models to generate intelligible, intuitive solution paths, including, \eg, medicine (where human doctors may need to make final decisions based on AI-generated diagnoses or treatment plans),
law (where human judges may need to rule in cases presented by AI lawyers),
computer use (where human users may hand off control over their computers to AI agents and \textit{vice versa}),
or mathematical reasoning (where human mathematicians may collaborate with AI models to prove new theorems).

As a concrete example from the math domain, consider a human collaborating with a reasoning language model to 
``Prove that the sum of the first $n$ natural numbers is $n(n+1)/2$.''
A large and well-trained reasoning model might have deep knowledge of advanced mathematical techniques and might wield them precisely; \eg, it might begin its solution by writing:
``Define the generating function for the sequence $a_k = k$: $A(x) = \sum_{k=1}^{\infty} kx^k$.''
Although this \textit{ansatz} is in principle correct and might be entirely sensible from the model's point of view, a human might not be able to follow along, which would complicate collaboration with, and verification by, human partners.
A more desirable solution might start by stating:
``By induction: For $n = 1$, we have $n(n+1)/2 = 1$.''

We ask: How might one incentivize language models to produce such intelligible solutions?
Potential ways forward include test-time approaches, such as system prompts describing the nature of intelligible, intuitive solutions, and training-time approaches, such as supervised finetuning on ground-truth intelligible solutions or RL with intelligibility rewards.
In practice, such approaches are, however, difficult to implement, as they require explicitly defining intelligibility \textit{a priori}---a hard-to-codify notion that may differ across settings, agents\slash users, and times.

We seek more viable methods for encouraging intelligible outputs without the need to explicitly define intelligibility, relying instead on the notion of \textit{handoff robustness,} which pragmatically and implicitly \textit{defines a partial solution as intelligible to another agent (a model or human) if that agent could continue the solution---at least for a few steps---without derailing it into failure.}
We operationalize this idea via \textit{tandem training,} in which two models---called \textit{senior} and \textit{junior}---take turns randomly during output generation, without coordinating.
The (typically weaker) junior model remains frozen, whereas the (typically stronger) senior model is trained%
\footnote{
Note the difference from model distillation \citep{hinton2015distilling,sanh2019distilbert}, where, in a reversal of roles, the stronger model is frozen while the weaker model is trained.
}
based on the quality of the output that the two models co-created.
In rollouts that concluded successfully despite the junior's participation, the senior acted in a way that enabled the junior to not make critical mistakes (or else the rollout would have failed), which, per our definition, means that the senior acted in a way that was intelligible to the junior.
Reinforcing the senior's behavior observed in successful rollouts thus achieves the dual objective of
making the senior more intelligible to the junior and keeping the senior's performance high.
The stochasticity of turn-taking not only provides a simple rule for when to switch between models, but also encourages handoff robustness and intelligible outputs in any situation, and prevents the senior model from acquiring tricks and reward hacks.

Tandem training can be viewed as a form of regularization:
by injecting noise into the training process, it encourages simpler, more generalizable behavior.
The approach is especially similar in spirit to dropout in neural\hyp network training \cite{srivastava2014dropout}, where neurons of the network being trained are randomly ``muted'' so the network learns not to over-rely on specific activation patterns.
Similarly, in tandem training, the senior model is randomly ``muted'' (and replaced by the ``noisier'' junior) so the senior learns not to over-rely on specific reasoning and speaking patterns.
Akin to other regularization methods, including dropout, noise injection (here via handoffs to the junior model) is performed only during training;
at test time, the tandem-trained senior model generates solutions on its own.

At a lower level, tandem training alternates between two phases:
(1) generating tandem rollouts, and
(2) updating the senior's policy based on them.

\xhdr{Tandem rollout generation}
To generate tandem rollouts, we devised a decoding algorithm where two language models $\Msen$ and $\Mjun$ work together to co-create an output.
The granularity of stochastic turn-taking is a design parameter that determines the atomic \textit{units} of text between which a handoff from one model to the other can occur, such as tokens, words, sentences, paragraphs, reasoning steps, etc.
The tandem decoder keeps both models in GPU memory.
Abstractly, the same input $x$ is fed to both models, but as the models may use different prompting modalities (\eg, language, system prompt, demonstrations, chat template), the concrete text sequences $x_\sen$ and $x_\jun$ seen by the two models may differ.
To co-create a shared response $y$, whenever a new token $y_{t+1}$ is to be generated to continue the partial response $y_{1:t}$, each model $m \in \{\sen, \jun\}$ independently samples a token $y^m_{t+1} \sim M_m(x_m \, y_{1:t})$ given the shared context.
Let $m_t$ be the currently active model (where $m_1$ is chosen randomly).
If appending $y^{m_t}_{t+1}$ to $y_{1:t}$ would begin a new unit (\eg, word or sentence), we toss a coin (we use $p=0.5$) to determine the new active model $m_{t+1}$; else, $m_{t+1} = m_t$ (since the current unit has not concluded yet).
Last, we extend the shared partial solution by $m_{t+1}$'s proposal:
$y_{1:t+1} = y_{1:t} \, y^{m_{t+1}}_{t+1}$.


\xhdr{Senior policy update}
In order to update the senior model based on tandem rollouts, tandem training can leverage any RL method for language modeling, including REINFORCE \citep{williams1992simple}, PPO \citep{schulman2017ppo}, GRPO \citep{shao2024deepseekmath}, etc., which perform gradient descent to maximize the expected reward of rollouts, where rewards may be obtained from programmatic verifiers, trained reward models, humans, etc.
We emphasize that tandem training does not require any explicit information about the differences between the senior and junior models, such as skill level, expected formatting, domain\hyp specific jargon, etc.;
updates are entirely based on the success of tandem rollouts.
This is important as the differences between senior and junior might be subtle, difficult to express, or unknown.

\section{Experimental setup}
\label{sec:Experimental setup}

To explore and evaluate the paradigm of tandem training for language models, we conduct experiments in the domain of mathematical reasoning, which offers multiple benefits as a first testing ground for tandem training.
First, researchers have shown that strong math reasoning models can be trained using RL without human supervision, relying solely on rewards obtained by automatically verifying answer correctness~\citep{kaliszyk2018reinforcement,wang2025oneshotrlvr,wen2025rlvr}.
In the absence of auxiliary rewards encouraging human-readable reasoning traces, such models have no incentive to reason in intelligible ways, which has already led to concerning outcomes, such as the ``poor readability and language mixing'' demonstrated by DeepSeek-R1-Zero~\citep{guo2025deepseek}.
There is an immediate need to prevent reasoning models from becoming unintelligible to humans, and tandem training holds promise to directly address this pressing challenge.

In addition to its practical relevance, math reasoning has the advantage that the wide difficulty range of math problems allows us to begin by working with smaller models.
Although tandem training is inspired by the challenge of making superhuman AI behave intelligibly to humans, we may begin with less capable model pairs.
Instead of considering a superhuman AI as senior and a human as junior, we may use a stronger, but still subhuman, model as senior and a weaker model as junior---shifting the performance levels downward while maintaining a capability differential.

We work with the GSM8K (Grade School Math 8K) benchmark and construct tandem pairs from differently trained or prompted variants of the \llama model family, as described next.%
\footnote{All code, models, and data are made publicly available at \url{https://github.com/epfl-dlab/lm-tandem-training}}

\subsection{Data}
\label{sec:Data}

GSM8K consists of 8,792 math word problems (7,473 for training, 1,319 for testing) that typically require two to eight steps of elementary arithmetic operations to solve~\citep{cobbe2021training}.
The dataset consists of English question--answer pairs, \eg,

{\small
\begin{list}{}{\setlength{\leftmargin}{1.6em}\setlength{\rightmargin}{1em}}
\item[Q:]
Julie is making Caesar salad for a family picnic. At the market, she spends \$8 on green lettuce and \$6 on red lettuce. If each type of lettuce costs \$2 per pound, how many total pounds of lettuce did she buy?
\item[A:]
The total cost of the green and red lettuce is \$8 + \$6 = \$\lgui 8+6=14\rgui 14.
Julie bought \$14 / \$2 = \lgui 14/2=7\rgui 7 pounds of lettuce.
\#\#\#\# 7
\end{list}
}

In the ground-truth answer, the final numerical solution is always given after ``\#\#\#\#''.
We also point out the special notation used for performing arithmetic: whenever an equality sign appears in the text (\eg, ``\$14 / \$2 =''), it is followed by a ``clean'' version (\eg, without dollar signs) of the preceding calculation, enclosed in double angle brackets (\lgui\rgui{}).
Presumably, this notation was introduced to facilitate parsing out arithmetic operations as ``programs''.
In our setting, we may consider it domain-specific ``jargon'' that a generalist model would not use by default.

\subsection{Models}
\label{sec:Models}

We construct tandem pairs from various language models derived from Llama-2-7b:%
\footnote{\url{https://hf.co/meta-llama/Llama-2-7b}}

\xhdr{Specialist model}%
\footnote{\url{https://hf.co/RedHatAI/Llama-2-7b-gsm8k}}
Obtained from Llama-2-7b via supervised finetuning on the training portion of GSM8K, this model demonstrates increased performance on GSM8K (39\% test accuracy, vs.\ 24\% for vanilla Llama-2-7b-chat).
It is prompted with the question only (no system prompt, no chat template) and, per its training data, uses \lgui\rgui{} jargon (\cf\ \Secref{sec:Data}) in its answers.

\xhdr{Base models}
Here we prompt the chat version of Llama-2-7b%
\footnote{\url{https://hf.co/meta-llama/Llama-2-7b-chat-hf}}
in one of five languages $L$ (English, German, French, Bulgarian, Serbian),
using a system prompt and two in-context question--answer demonstrations to explain the task and the expected output format and language (see \Appref{app:prompts}), followed by the GSM8K question.
The entire prompt is provided in language $L$ (questions were translated ahead of time using GPT-4.1-mini),
which effectively switches the model's default language from English to $L$.
As the base models were not specifically tuned for GSM8K, they do not produce \lgui\rgui{} jargon.
GSM8K test accuracy ranges between 12\% (Serbian) and 24\% (English).

\begin{figure*}[t]
\begin{center}
\includegraphics[width=\textwidth]{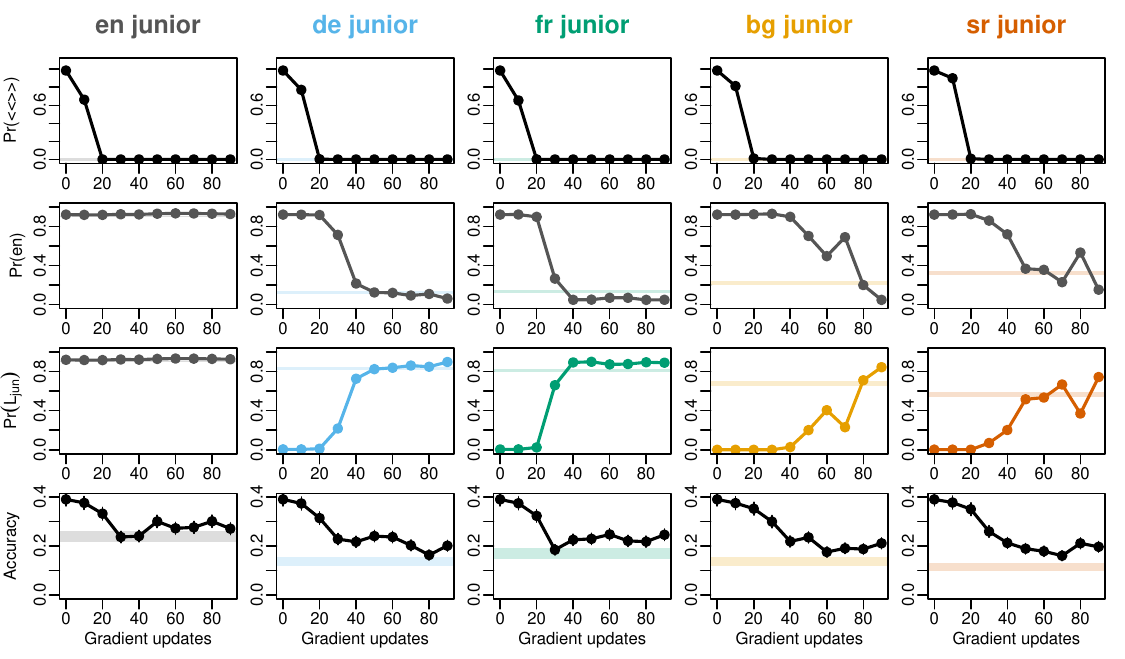}
\end{center}
\caption{
\textit{\textbf{Column~1:}}
Results for \textit{skill disparity} setting, where tandem-trained senior is GSM8K specialist, and frozen junior is \llama prompted in English.
\textit{\textbf{Columns~2--4:}}
Results for \textit{skill and language disparity} setting, where junior is \llama prompted in non-English languages (one per column: German, French, Bulgarian, Serbian).
\textit{\textbf{Row~1:}}
Use of notational jargon (\lgui\rgui).
\textit{\textbf{Row~2:}}
Use of linguistic jargon (English).
\textit{\textbf{Row~3:}}
Use of junior language.
\textit{\textbf{Row~4:}}
Accuracy.
\textit{\textbf{Curves:}}
Tandem-trained senior (with 95\% CIs).
\textit{\textbf{Shaded bands:}}
Frozen junior (95\% CIs).
\textit{\textbf{Takeaway:}}
Tandem training reduces senior model jargon while keeping senior accuracy above junior level.
}
\label{fig:results_skill_disparity}
\end{figure*}

\subsection{Training and testing}
\label{sec:Training and testing}

We consider three settings, each of which combines the above models in a way that allows us to observe whether tandem training produces the desired effects (input languages in parentheses):

\begin{enumerate}
\denselist
\item \textbf{Skill disparity:}
$\Msen =$ specialist (English);
$\Mjun =$ base (English);
jargon $=$ \{\lgui\rgui{}\}
\item \textbf{Skill \& language disparity:}
$\Msen =$ specialist (English);
$\Mjun =$ base ($\Ljun \neq$ English);
jargon $=$ \{\lgui\rgui{}, English\}
\item \textbf{Language disparity:}
$\Msen =$ base ($\Lsen$);
$\Mjun =$ base ($\Ljun \neq \Lsen$);
jargon $= \{\Lsen\}$
\end{enumerate}


In the \textit{skill disparity} setting, we should expect successful tandem training of the senior model to decrease usage of the GSM8K-specific \lgui\rgui{} jargon, while maintaining task accuracy well above that of the junior model.
In the \textit{skill and language disparity} setting, the senior additionally differs from the junior with respect to its input language, so in addition to \lgui\rgui{} jargon, we should expect successful tandem training to also make the senior respond in the junior's input language $\Ljun$, rather than in English (the senior's input language).
Finally, in the \textit{language disparity} setting, neither the senior nor the junior was specifically tuned for GSM8K, so \lgui\rgui{} jargon plays no role here. Instead, the senior's input language $\Lsen$ can be considered jargon, and we should expect successful tandem training to make the senior respond in the junior's, rather than its own, input language.

In all settings, we tandem-train the senior for one epoch on the GSM8K training portion and evaluate it (on its own, without the junior) on the testing portion.
Training is done using a variant of the REINFORCE~\citep{williams1992simple} RL algorithm with binary rewards, where the following steps are iterated:
(1)~Given a batch of input questions, sample a set of (in our case, two) tandem rollouts for each question (\cf\ \Secref{sec:Method}).
(2)~Perform an SGD update to increase the log-likelihood of the correct rollouts, while discarding incorrect rollouts.

When computing the log-likelihood of a rollout, one may also mask tokens produced by the junior, as they were not produced by the (senior) model being trained, but for simplicity we did not perform masking in the main experiments reported here.
Turn-taking in tandem rollouts was carried out uniformly at random at the word level (\cf\ \Secref{sec:Method}).
The senior model was tuned using low-rank adapters on all linear layers~\citep{hu2022lora}.
For hyperparameters, see \Appref{app:Hyperparameters}.

We emphasize that our goal is not to develop new RL algorithms, but to showcase the novel tandem training paradigm, which can, in principle, be used with any RL algorithm.
Hence, it is advantageous if tandem training works even with a simple RL algorithm, such as the one presented here.
(See \Appref{app:Supplemental results} for an exploration of more complex variants, allowing for soft-masking of junior tokens and for not discarding incorrect rollouts.)

\section{Results}
\label{sec:Results}

Since the two-fold goal of tandem training is for the trained senior model to stop using jargon while keeping task accuracy high, we track accuracy as well as jargon over the course of training.
We do so by storing checkpoints after every 10 gradient updates and using them to generate senior answers for the GSM8K test set.
\textbf{Accuracy} is measured as the fraction of answers that provide the correct numerical solution.
Jargon is measured in two ways:
\textbf{notational jargon} is the fraction of answers that contain \lgui{} or \rgui{}, whereas
\textbf{linguistic jargon} is the probability of the senior's input language $\Lsen$ in its generated output, according to a language identification method based on a fastText model (see \Appref{app:Language identification}).
Next, we discuss results for each of the three settings described in \Secref{sec:Training and testing}.

\xhdr{Skill disparity}
The leftmost column of \Figref{fig:results_skill_disparity} shows results for the \textit{skill disparity} setting, where, during tandem training, an (English) GSM8K-specialist senior is paired with a base-\llama junior prompted in English.
We observe that, whereas before tandem training the senior uses notational jargon (\lgui\rgui) in nearly every answer (99\%),
it entirely stops doing so (0\%) within 20 gradient updates.%
\footnote{
\label{fn:jargon for RL-only}
To confirm that jargon indeed disappears due to tandem training, rather than simply due to RL, we isolated the effects of tandem training and RL in a supplementary experiment where we trained the senior using the same RL method, but on rollouts produced by the senior alone.
In this setup, jargon remains at 99\% for all checkpoints, confirming tandem training, rather than mere RL, as the cause for vanishing jargon.
}
At this point, accuracy is 33\%, slightly down from the 39\% achieved by the senior before tandem training, and considerably higher than the 24\% achieved by the frozen junior.
Thereafter, accuracy first decreases and then remains stable and largely above junior level.
We discuss and mitigate the accuracy decrease in the final paragraph of this section.



\begin{figure*}[t]
\centering
\includegraphics[width=0.97\linewidth]{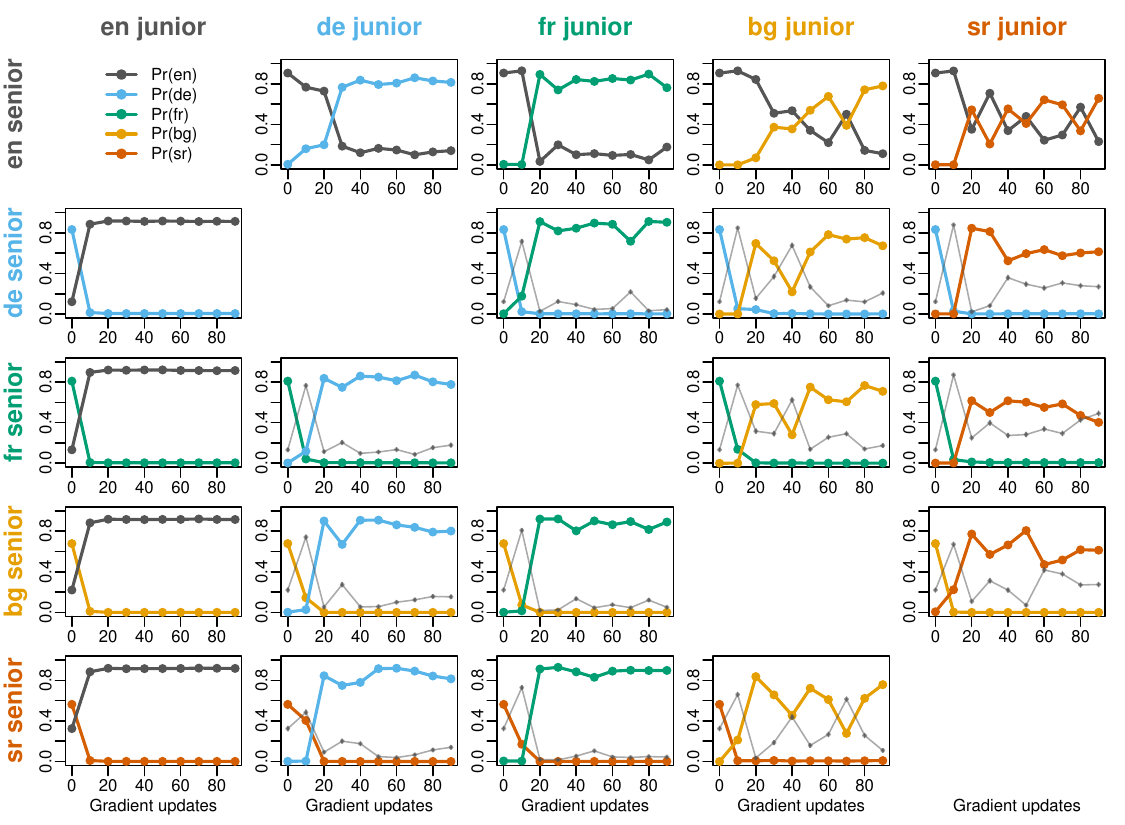}
\caption{
Results for \textit{language disparity} setting,
where tandem-trained senior \textit{\textbf{(rows)}} and frozen junior \textit{\textbf{(columns)}} are \llama models prompted in different languages.
Plots show senior's language use when applying checkpoints from tandem training to GSM8K test data.
(Senior and junior accuracy shown in \Figref{fig:results_lang_disparity__accuracy}.)
\textit{\textbf{Takeaway:}}
Tandem training reduces senior jargon (use of senior's input language) without driving accuracy below junior level.
}
\label{fig:results_lang_disparity__langprob}
\end{figure*}

\xhdr{Skill and language disparity}
The four right columns of \Figref{fig:results_skill_disparity} show results for the \textit{skill and language disparity} setting, where tandem training pairs the (English) GSM8K-specialist junior with a base-\llama junior prompted in a non-English language $\Ljun$.
Notational jargon (\lgui\rgui) vanishes as quickly as in the above\hyp discussed skill disparity setting (within 20 gradient updates).
Linguistic jargon also disappears as a consequence of tandem training: for three of the four junior languages $\Ljun$, the senior has entirely abandoned English in favor of $\Ljun$ within 50 gradient updates; for the fourth junior language (Bulgarian), within 80 updates.
At the same time, senior accuracy remains significantly above junior level throughout training.
Overall, tandem training again has the desired effect: to make senior jargon disappear while keeping accuracy above junior level.

\begin{figure*}[t]
\includegraphics[width=0.97\linewidth]{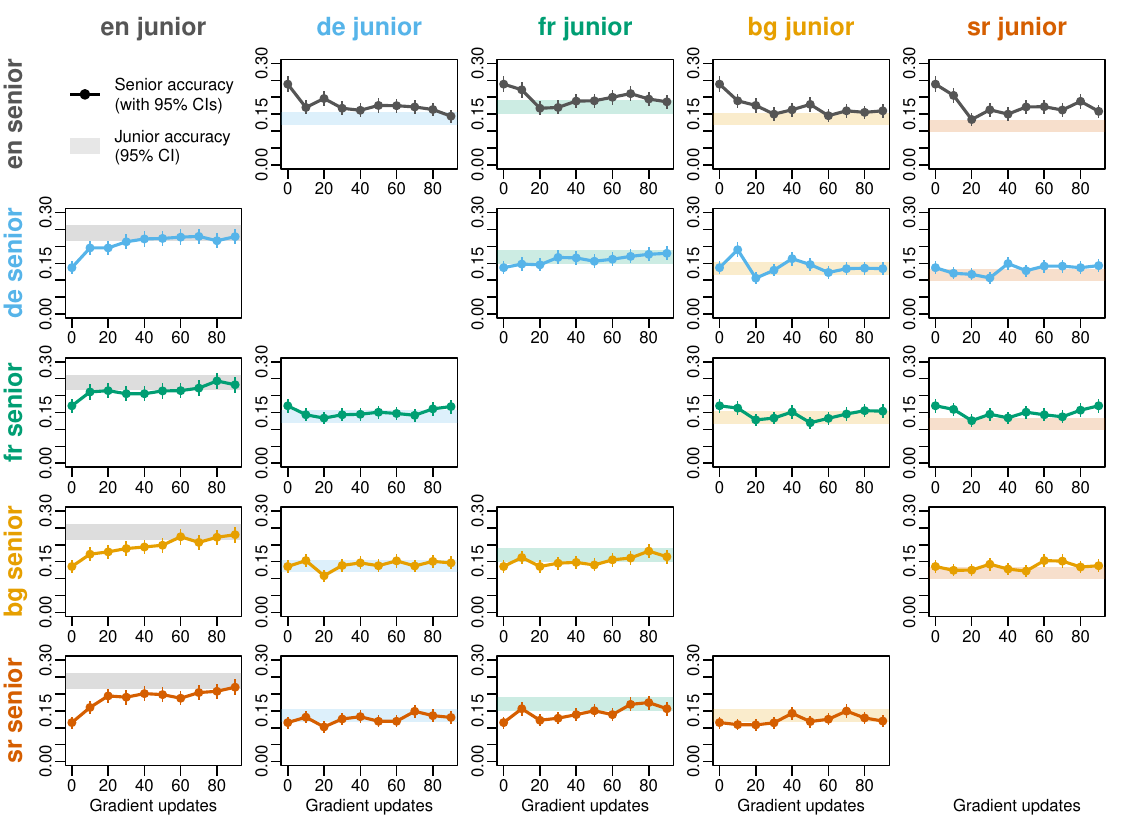}
\centering
\caption{
Senior and junior accuracy for \textit{language disparity} setting,
where tandem-trained senior \textit{\textbf{(rows)}} and frozen junior \textit{\textbf{(columns)}} are \llama models prompted in different languages.
}
\label{fig:results_lang_disparity__accuracy}
\end{figure*}

\xhdr{Language disparity}
Finally, we consider the \textit{language disparity} setting, where both the senior and junior are base-\llama models, but prompted in different languages $\Lsen$ and $\Ljun$, respectively.
Here, the senior's input language $\Lsen$ is considered jargon.
As observed in \Figref{fig:results_lang_disparity__langprob}, tandem training leads the senior to abandon its jargon fast and adopt the junior's language instead, typically within 20 gradient updates.
In terms of accuracy (\Figref{fig:results_lang_disparity__accuracy}), we find that,
when the junior beats the senior before training, the senior catches up;
when they do equally well before training, it remains this way;
and when the senior beats the junior before training, senior accuracy tends to decrease (see discussion in next paragraph), but never below junior accuracy.
The tenor is again that tandem training makes jargon disappear without driving accuracy below junior level.
Note in \Figref{fig:results_lang_disparity__langprob} that the switch from $\Lsen$ to $\Ljun$ is never direct, but always passes through an intermediate phase (around gradient update 10) where the senior outputs English---which is neither $\Lsen$ nor $\Ljun$.
In other words, English appears as a transitory \textit{lingua franca}.
Inspection of tandem rollouts from training shows that, in early training, rollouts often begin with a mix of $\Lsen$ and $\Ljun$, which seems to confuse the models and lead them to switch to English amidst rollouts.
As such English completions tend to be more successful, the trained senior adopts this behavior, but the frozen junior maintains its tendency to produce $\Ljun$, especially early on in rollouts.
This, in turn, lets the senior eventually abandon English in favor of $\Ljun$.

\xhdr{Mitigating accuracy decrease}
In the above results, senior accuracy tends to decrease towards junior accuracy over the course of tandem training.
We hypothesize that the decrease is due to a gradual distribution shift:
easier problems are more likely to be solved correctly, and since our simple RL method discards incorrect rollouts, the senior adapts to easier problems over the course of training. However, since the test set still follows the overall difficulty distribution, test accuracy suffers.
This hypothesis is supported by a supplementary experiment where RL was performed with rollouts from the senior alone, rather than tandem rollouts (\cf\ footnote~\ref{fn:jargon for RL-only}): here, too, test accuracy decreased with more training (from 40\% to 34\%), pointing to the RL method, rather than tandem rollouts, as the cause of accuracy deterioration.
We thus experimented with two modifications to the RL method:
(1)~instead of discarding incorrect rollouts, include them in the log\hyp likelihood objective with negative weight $c<0$;
(2)~instead of weighing senior and junior tokens equally in the objective, ``soft-mask'' junior tokens via a multiplicative factor $j<1$.
Modification~1 enables learning from failure; modification~2 lets the senior focus more on its own than on the junior's behavior.
(Note that the simpler RL method used in the main experiments above corresponds to $c=0, j=1$.)
In \Appref{app:Supplemental results} we show that, by optimizing the hyperparameters $(c,j)$ on a validation set, the accuracy decrease is strongly mitigated, while jargon still vanishes entirely---the best of both worlds.

\section{Discussion}
\label{sec:Discussion}

\xhdr{Result summary}
Across all three settings, tandem training rapidly suppresses ``jargon'', indicating the senior model learning to adapt to become compatible with the junior model, while maintaining above\hyp junior accuracy.
This is the core promise of tandem training: we can incentivize an AI model to adapt its behavior and actions to be compatible with a given weaker collaborator without an overly negative impact on its capabilities.
Taken together, the results show that handoff robustness, our method of encouraging intelligibility, can be induced directly inside the RL trajectory by randomizing which partner generates next, creating pressure for the senior to produce continuable reasoning traces rather than idiosyncratic ones.

\xhdr{Learning signals}
Our current sequence\hyp level objective attributes rewards to entire rollouts, which is simple and effective but coarse. Three complementary refinements target sharper, lower\hyp variance credit signals. First, masking junior-authored tokens during likelihood and policy-gradient computations can prevent spurious credit from leaking onto text that the trainable senior did not produce, aligning gradients with the senior's actual contributions. Second, learning from negative examples can leverage failed rollouts as informative counterfactuals: contrastive or penalty terms could down-weight the specific trajectories and specific segments most predictive of handoff failure. Third, token- and span-level preference learning~\citep{deng2024flow} can localize blame and credit to the precise tokens where handoffs break, rather than pooling signal uniformly across the trajectory. 
Our work shows that tandem training is effective even when combined with a simple RL method, and we expect the benefits to increase further with more sophisticated RL methods.
This expectation is supported by the initial experiments laid out at the end of \Secref{sec:Results} and in \Appref{app:Supplemental results}, which show that the first two refinements listed above improve senior task accuracy.

\xhdr{Extending the framework}
We present an atomic setting with a frozen junior and a trained senior, but there are many natural extensions of this basic tandem training framework. Co-adaptation would be possible by training both models, either simultaneously or by alternating which is trained and which is frozen, perhaps with additional anchoring constraints to prevent private codes from emerging. The ``junior model'' role can be expanded to design and control the type of compatibility we want to produce in the senior model; one could imagine varying junior competence, style, language, tool use, etc., to regularize the senior toward broadly intelligible behavior. An automated framework could maintain a pool of juniors that are swapped in via a bandit algorithm or curriculum policy where juniors that expose failure modes are prioritized and juniors that the senior has already mastered are gradually retired. It is also possible to apply tandem training earlier in the stack (\eg, late pretraining or supervised finetuning) to encourage intelligibility as a more fundamental attribute. Finally, we limited our attention to fixed i.i.d.\ handoffs, which could likely be improved with a handoff schedule that optimizes robustness.

\section{Conclusion}
\label{sec:Conclusion}

This work introduces tandem training for language models, a novel RL paradigm that operationalizes and encourages intelligibility via handoff robustness.
The approach is lightweight, architecture\hyp agnostic, and complementary to existing RL pipelines.
Our results show that by making intelligibility a prerequisite for reward, tandem training aligns AI models toward behavior that partners can pick up, audit, and extend, with benefits for scalable oversight and practical collaboration.
We see great promise in harnessing tandem training in multi-agent systems and human--AI collaboration.


\section{Limitations}
\label{sec:Limitations}


Our principal goal is to investigate the viability of tandem training for encouraging language models to produce more intuitive and intelligible output.
The settings and methods we consider in this first paper are hence limited in scope:

\begin{itemize}
\item We consider a single application domain (mathematical reasoning) and evaluate on a single benchmark (GSM8K). An important avenue for future work is to investigate whether the paradigm works equally well in other domains, and what makes a domain more or less amenable to tandem training.
\item Although tandem training can in principle be combined with any RL algorithm, we tested it only with REINFORCE with binary rewards. While future work should improve the RL method (\cf\ discussion in \Secref{sec:Discussion}), we consider it an advantage that tandem training works even with simple RL, and we expect it to be even more effective when combined with more advanced RL.
\item The tandem decoder implementation described in \Secref{sec:Method} assumes that both models (junior and senior) use the same tokenizer. An extension to different tokenizers would be straightforward.
\end{itemize}

\section{Ethical considerations}
This work investigates tandem training as a way to operationalize intelligibility by requiring that a strong model's partial solutions be continuable by a weaker collaborator under randomized handoffs during RL. Our experiments use public math word-problem data (GSM8K) and off-the-shelf Llama-2-7b variants; no human subjects or personal data are involved.

\xhdr{Potential benefits and misuse} By design, tandem training increases handoff robustness and reduces jargon, which can strengthen auditability and scalable oversight by weaker agents. However, improved continuability could be misapplied to coordinate more effectively with colluding AIs or to make reasoning traces more persuasive in undesirable contexts. To mitigate this, we recommend anchoring mechanisms, \eg, human\hyp language constraints, verifiers, and audits, especially when both partners are trained, to deter private codes and preserve human understandability.

\section{Reproducibility statement}
Experiments were run on machines with one or two Nvidia A100 GPUs with 80~GB of memory each.
Data, models, training and testing procedure, and evaluation methodology are described in the main text.
Hyperparameters, prompts, and further method details are listed as appendices.
Model artifacts used to instantiate seniors and juniors are referenced in the main text.
We release all code, models, and data required to reproduce our results.


\section*{Acknowledgments}
This work was performed when Robert West and Ashton Anderson spent their sabbaticals as Visiting Researchers with the AI Frontiers group at Microsoft Research.
Thanks to all our Microsoft colleagues for their thoughtful feedback on this work (in particular Saleema Amershi, Siddhartha Sen, and Tim Davidson) and for their warm welcome in Redmond and New York!
We also gratefully acknowledge the help of Raghav Singhal and Difan Jiao, who ran additional experiments during the review phase (\Appref{app:Supplemental results}).

\bibliography{bibliography}


\appendix

\section{\llama prompts}
\label{app:prompts}

Llama-2-7b-chat was prompted for GSM8K with a system prompt, followed by two question--answer demonstrations, followed by the input question.

\subsection{System prompt}

\begin{quote}
\textit{%
You are a math reasoning assistant specialized in solving grade school math word problems. Your task is to provide a clear, step-by-step explanation of the solution and, importantly, to output the final numerical answer on a separate line in exactly the following format:\\
``\#\#\#\# <final answer>''\\
where <final answer> is the numerical result (without any units or extra symbols). Do not include any additional text or formatting on that final line, just a single number.
}
\end{quote}

When the model was used in a non-English language $L$, the above system prompt was extended by the following instruction and translated into $L$:
\begin{quote}
    \textit{Respond exclusively in $L$. Never use English or any other language. If you use the English language or any other language, the answer is invalid. Every part of the response must be entirely $L$.}
\end{quote}

\subsection{Demonstrations}

Showing demonstrations serves to exemplify the expected output format and language described in the system prompt.
The following two question--answer pairs wer used as demonstrations:

\begin{quote}
\textsc{[user]} \textit{Natalia sold clips to 48 of her friends in April, and then she sold half as many clips in May. How many clips did Natalia sell altogether in April and May?}

\textsc{[assistant]} \textit{Natalia sold 48/2 = 24 clips in May. Natalia sold 48+24 = 72 clips altogether in April and May. \#\#\#\#~72}

\textsc{[user]} \textit{Betty is saving money for a new wallet which costs \$100. Betty has only half of the money she needs. Her parents decided to give her \$15 for that purpose, and her grandparents twice as much as her parents. How much more money does Betty need to buy the wallet?}

\textsc{[assistant]} \textit{In the beginning, Betty has only 100 / 2 = \$50. Betty's grandparents gave her 15 * 2 = \$30. This means, Betty needs 100 - 50 - 30 - 15 = \$5 more. \#\#\#\# 5}
\end{quote}

When the model was used in a non-English language $L$, the demonstrations were also translated to $L$, with the following exception:
during development, we found that \llama is less likely to fall back to English (its majority training language), especially during mixed-language tandem rollouts, when one demonstration has a question in English and an answer in $L$, emphasizing that the model should always revert to $L$.
We hence always keep the first demonstration's question (but not its response) in English.

\section{Hyperparameters}
\label{app:Hyperparameters}


\begin{tabular}{ll}
Effective batch size        & 152 or 156 \\
Learning rate               & $10^{-4}$ \\
Training temperature        & 0.7 \\
Testing temperature         & 0 \\
Max output length           & 256 \\  
Max input + output length   & 512 \\
Training rollouts per prompt& 2 \\
Torch dtype                 & \texttt{bfloat16} \\
QLoRA base model quantization  & 4 bits \\
QLoRA rank $r$              & 16 \\
QLoRA scaling factor $\alpha$ & 16 \\
QLoRA target modules        & All linear \\
\end{tabular}

\bigskip

With rollout diversity in mind, when performing two rollouts per prompt, the second rollout's coin-toss sequence was an inverted version (\ie, with heads and tails swapped) of the first rollout's coin-toss sequence.

``All linear'' QLoRA target models:
\texttt{q\_proj},
\texttt{k\_proj},
\texttt{v\_proj},
\texttt{o\_proj},
\texttt{gate\_proj},
\texttt{up\_proj},
\texttt{down\_proj}.


\section{Language identification}
\label{app:Language identification}

In order to infer the languages present in the model outputs, we rely on the fastText \cite{joulin2016fasttext,joulin2016bag} \texttt{lid.176.bin} language identification model.%
\footnote{\url{https://fasttext.cc/docs/en/language-identification.html}}
Since this model was not trained for mixed-language text (as often produced during tandem training), we do not apply the model directly to the entire text, but use a sliding-window approach instead, as short windows generally contain less language mixing.
First, the input text is cleaned of digits and symbols
(\#, =, +, -, *, <, >, $\times$, $\div$, \$)
and split into tokens.
It is then divided into overlapping windows (``shingles'') of eight tokens each.
Each shingle is passed to the fastText language identification model, which returns a probability distribution over possible languages.
Finally, these per-shingle distributions are averaged across all shingles to produce a global language distribution for the text.

\section{AI assistants statement}
\label{app:AI Assistants}
We used AI assistants like Copilot and GPT models to help us with writing the tandem training code and to assist with writing.

\section{Supplemental results}
\label{app:Supplemental results}

\begin{figure*}[t]
  \begin{subfigure}[t]{0.5\linewidth}
    \centering
    \includegraphics[width=\linewidth]{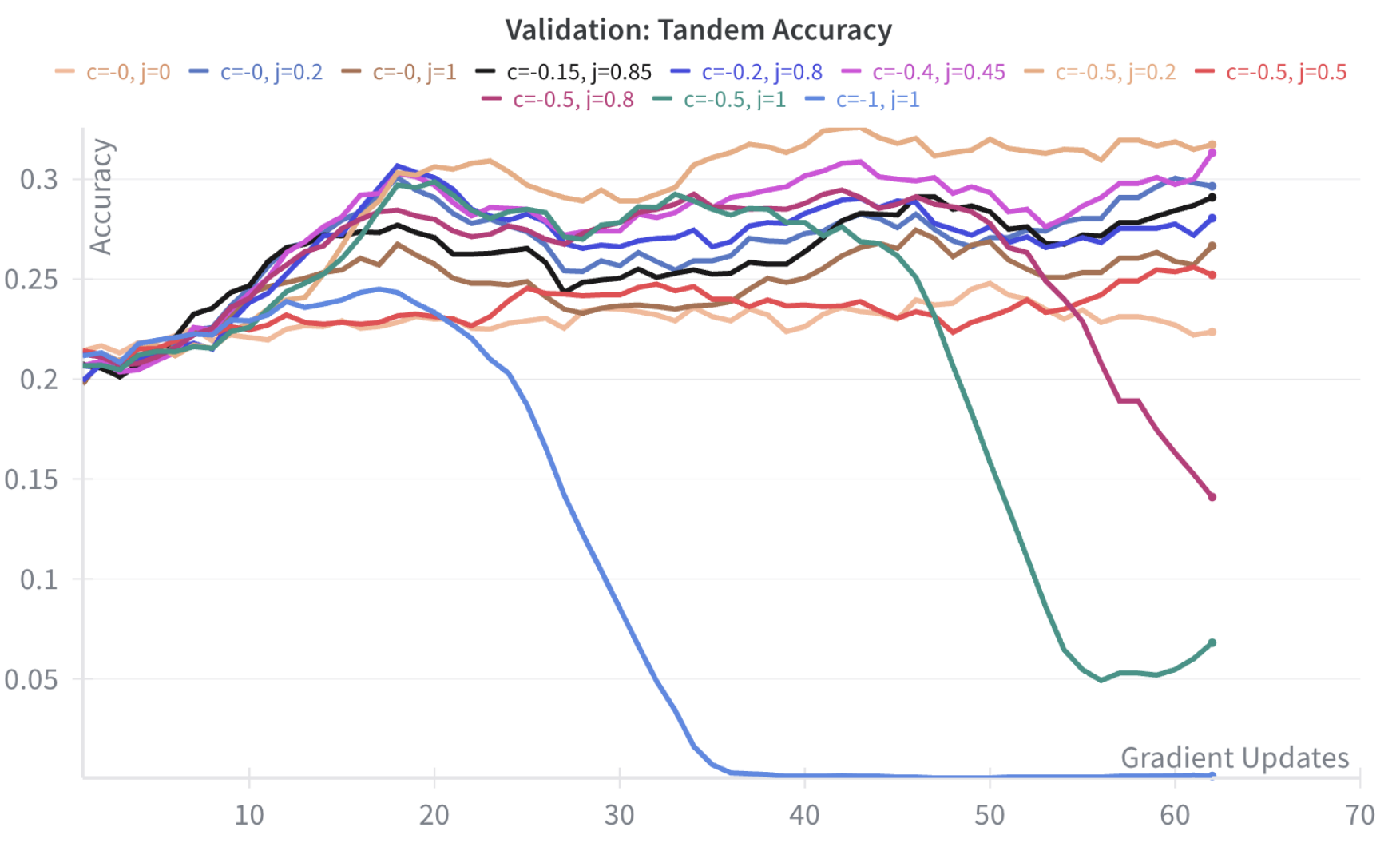}
    \caption{Validation accuracy: \textit{skill disparity} setting.}
    \label{fig:ablations_val_en}
  \end{subfigure}
  \hfill
  \begin{subfigure}[t]{0.5\linewidth}
    \centering
    \includegraphics[width=\linewidth]{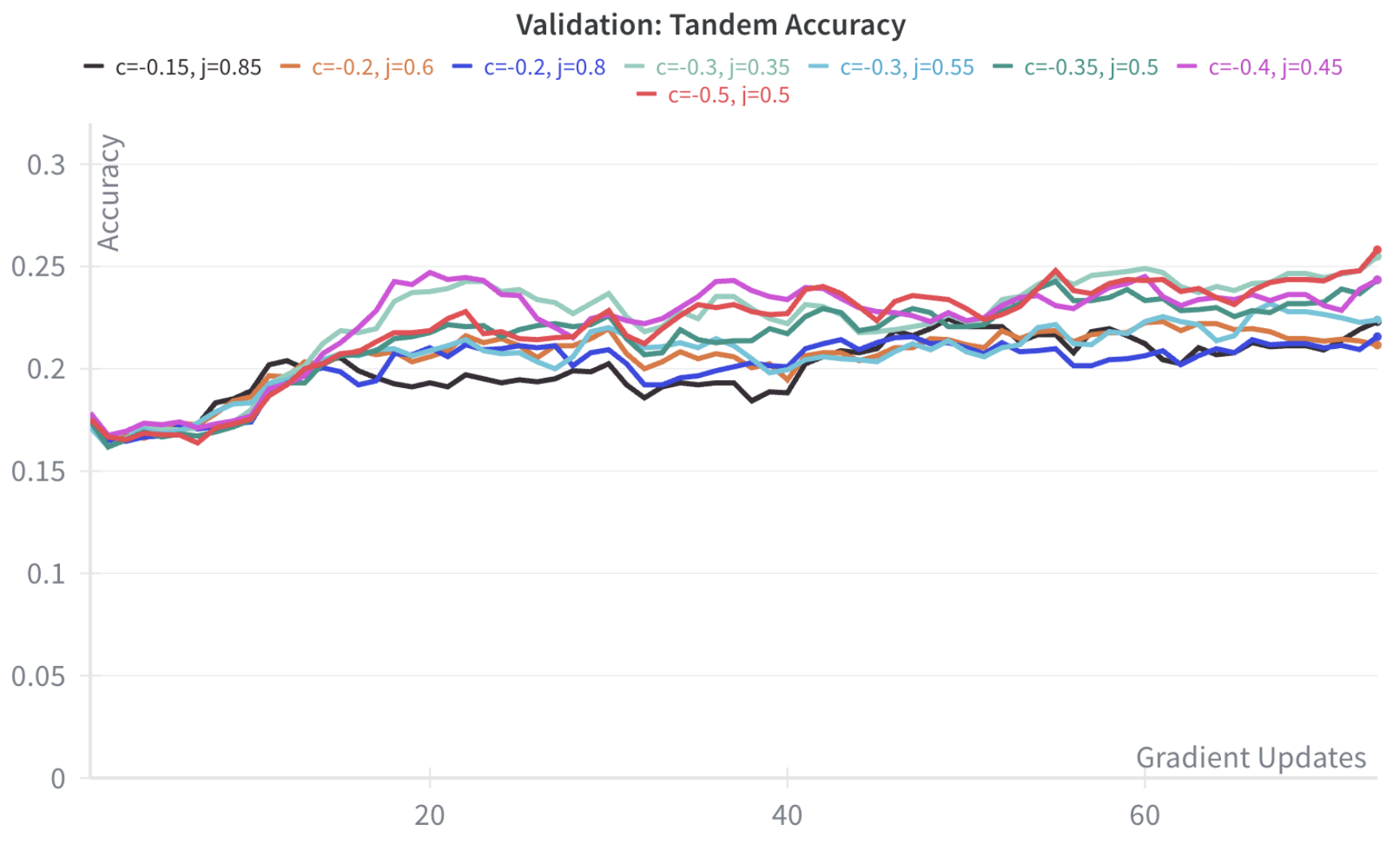}
    \caption{Validation accuracy: \textit{skill and language disparity} setting.}
    \label{fig:ablations_val_fr}
  \end{subfigure}

  \vspace{4mm}
  \begin{subfigure}[t]{\linewidth}
    \centering
    \includegraphics[width=\linewidth]{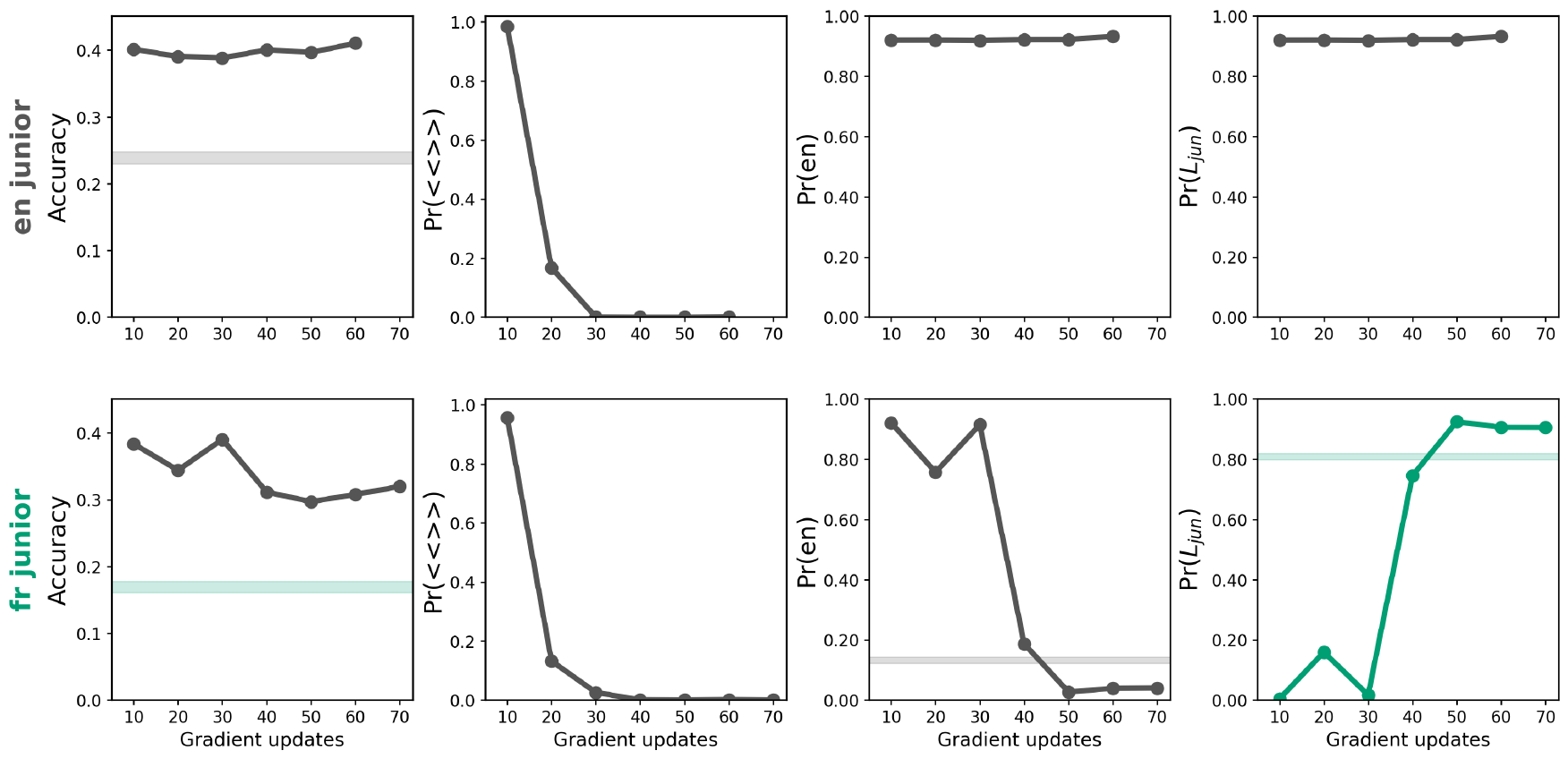}
    \caption{Test results for \textit{skill disparity} setting \textbf{\textit{(top)}} and \textit{skill and language disparity} setting \textbf{\textit{(bottom)}}.}
    \label{fig:ablations_test}
  \end{subfigure}
\caption{
Results for tuning the hyperparameters $(c,j)$ (defined in \Appref{app:Supplemental results}) of the RL algorithm.
\textbf{(a)} Validation accuracy of the tandem model for a range of hyperparameter choices, in the skill disparity setting.
\textbf{(b)} \textit{Idem,} in the skill and language disparity setting.
\textbf{(c)} Test results achieved by the senior models selected based on validation accuracy.
In comparison to \Figref{fig:results_skill_disparity} (columns 1 and~3), GSM8K accuracy remains higher, while jargon (\lgui\rgui{} in the case of skill disparity, \lgui\rgui{} and English in the case of skill and language disparity) still disappears.
}
\label{fig:ablations}
\end{figure*}

While the main experiments of \Secref{sec:Results} used a variant of REINFORCE with binary rewards (see \Secref{sec:Training and testing}), we also explored the following generalization:
\begin{enumerate}
    \item Instead of discarding incorrect rollouts, include them in the loss with negative weight $c<0$ (whereas correct rollouts are included with positive weight~1).
    \item Instead of weighing senior and junior tokens equally in the loss, ``soft-mask'' junior tokens via a multiplicative factor $j<1$.
\end{enumerate}
Modification~1 lets the senior learn from failure, whereas modification~2 lets the senior focus more on its own than on the junior's behavior.
The simpler RL method used in the main experiments corresponds to $c=0, j=1$.

We validated a range of $(c,j)$ combinations in order to investigate whether a judicious choice of these hyperparameters can mitigate or prevent the task-accuracy deterioration observed in \Secref{sec:Results} while still teaching the senior to avoid jargon.
Importantly, we should not use any predefined notion of jargon in order to choose hyperparameters.
One key reason for choosing GSM8K as a domain was precisely that it offers an exactly measurable notion of jargon (\lgui\rgui), which facilitates studying the effects of tandem training; but in real-world settings we generally do not know in what aspects---including jargon---senior and junior might differ.

We hence chose $(c,j)$ without referring to jargon and measured jargon only after the fact, to determine whether the corresponding senior model with the chosen hyperparameters meets the criteria of reducing jargon while maintaining high GSM8K task accuracy.
Concretely, for each $(c,j)$, we tandem-trained a senior model for one epoch on a random sample containing 90\% of the GSM8K training portion and validated the GSM8K task accuracy of the tandem (consisting of the trained senior and the frozen junior) on the remaining 10\%.
After training, we chose the $(c,j)$ for which the tandem had maximum validation accuracy on average over the last few training checkpoints. Here, the rationale is that, in order for a tandem---where senior and junior take random turns during rollouts---to have high GSM8K accuracy, the senior must have learned to adapt to the junior, which we take as a heuristic for choosing a suitable senior.

In these preliminary experiments, we restricted ourselves to the \textit{skill disparity} setting and to the \textit{skill and language disparity} setting with a French junior.
The training runs used the same hyperparameter values listed in \Appref{app:Hyperparameters}, with the exception of effective batch size, which, per GPU capacity, was set to 160 and 152, respectively.
The tandem model's validation accuracies are shown in \Figref{fig:ablations_val_en} and \ref{fig:ablations_val_fr}, respectively, leading us to choose
$(c=-0.5, j=0.2)$ for the skill disparity setting and
$(c=-0.3, j=0.35)$ for the skill and language disparity setting.
Evaluating the chosen seniors (on their own, not in a tandem) on the test portion of GSM8K (\Figref{fig:ablations_test}) shows that tandem training makes them stop using jargon (\lgui\rgui{}, and additionally English in the skill and language disparity setting) while keeping task accuracy considerably higher than in the main experiments:
whereas in the main experiments, task accuracy in the skill disparity setting noticeably dropped from its initial 39\% and stabilized around 30\% (\Figref{fig:results_skill_disparity}, first column), here it remained at around 40\% (\Figref{fig:ablations_test}, top row);
and whereas in the skill and language disparity setting with a French junior 
 (\Figref{fig:results_skill_disparity}, third column) it stabilized between 22\% and 24\%, here it remained above 30\% (\Figref{fig:ablations_test}, bottom row).

Taken together, these supplemental results show that a slightly more sophisticated variant of the RL method of \Secref{sec:Training and testing} more closely achieves the key promise of tandem training: to make the senior more intelligible without sacrificing performance.
We thus see exploring even more advanced RL algorithms as a promising direction for future work.

\end{document}